\documentclass[11pt,a4paper]{article}

\usepackage[table,dvipsnames]{xcolor}

\usepackage[hyperref]{acl2020}
\usepackage{times}
\usepackage{latexsym}

\usepackage{times}
\usepackage{latexsym}

\usepackage{url}
\usepackage{amsmath}
\usepackage{bm}
\usepackage{booktabs}
\usepackage{multirow}
\usepackage{bm}
\usepackage{dblfloatfix}    
\usepackage{capt-of}
\usepackage{amssymb}
\usepackage{mathabx}
\usepackage [autostyle, english = american]{csquotes}
\usepackage{tabularx}
\usepackage{subcaption}
\usepackage{verbatim}
\usepackage{graphicx}
 \usepackage{lipsum}
 \usepackage{caption}
\usepackage{subcaption}
\usepackage{csquotes}
\usepackage{tcolorbox,varwidth}

\usepackage{pifont}

\usepackage{float}

\usepackage{adjustbox}
\usepackage{array}
\usepackage{booktabs}
\usepackage{multirow}
\usepackage{enumitem}

\usepackage{algorithm}
\usepackage[noend]{algpseudocode}
\newcommand\sbullet[1][.5]{\mathbin{\vcenter{\hbox{\scalebox{#1}{$\bullet$}}}}}

\newcolumntype{R}[2]{%
    >{\adjustbox{angle=#1,lap=\width-(#2)}\bgroup}%
    l%
    <{\egroup}%
}

\definecolor{mygray}{gray}{0.3}
\definecolor{tablegray}{gray}{0.9}
   
\definecolor{colorzero}{HTML}{0924F5}
\definecolor{colorone}{HTML}{B02442}
\definecolor{colortwo}{HTML}{F08633}

\aclfinalcopy 
\def\aclpaperid{***} 

\title{The Unreasonable Volatility of \\Neural Machine Translation Models}

\author{\stepcounter{footnote}Marzieh Fadaee\Thanks{Marzieh is now affiliated with Zeta Alpha Vector.} \\
Informatics Institute \\
University of Amsterdam \\
  \texttt{marzieh.f@gmail.com} \\\And
 Christof Monz \\
Informatics Institute \\
University of Amsterdam \\
  \texttt{c.monz@uva.nl} \\}

\date{}

\begin{document}
\maketitle
\begin{abstract}
Recent works have shown that while Neural Machine Translation (NMT) models achieve impressive performance, questions about understanding the behaviour of these models remain unanswered.
We investigate unexpected volatility of NMT models where the input is semantically and syntactically correct.
We discover that with trivial modifications of source sentences, we can identify cases where \textit{unexpected changes} happen in the translation and in the worst case lead to mistranslations. 
This volatile behaviour of translating extremely similar sentences in surprisingly different ways highlights the underlying generalization problem of current NMT models. 
We find that both RNN and Transformer models display volatile behaviour in $26\%$ and $19\%$ of sentence variations, respectively.
\end{abstract}

\section{Introduction}

The performance of Neural Machine Translation (NMT) models has dramatically improved 
in recent years, and with sufficient and clean data these models outperform more traditional models.
Challenges when sufficient data is not available include translations of rare words \citep{W18-2712} and idiomatic phrases \citep{L18-1148} as well as domain mismatches between training and testing \citep{koehn2017six,W18-2709}.

\begin{table}[ht]
\footnotesize
\centering
\begin{tabularx}{0.97\columnwidth}{lll}
\multicolumn{3}{l}{\textcolor{mygray}{Source:} \textit{Ich bin \underline{\ \ \ \ \ \ }\ \raisebox{-1.5ex}{\tiny{\fbox{$1$}}} \textbf{erleichtert} und \underline{\ \ \ \ \ \ }\ \raisebox{-1.5ex}{\tiny{\fbox{$2$}}} bescheiden.}}\\
\hline
\\[-0.5ex]
\fbox{$1$} &  \fbox{$2$} & \textcolor{mygray}{NMT output} \\
\hline
  $\phi$ & $\phi$   & I am \underline{easier} and modest.  \\ 
  $\phi$ & \textit{sehr} $ ^{\ddagger}$&  I am \textbf{relieved} and very modest. \\ 
  \textit{sehr} & $\phi$ & I am {very} much \underline{easier} and modest.  \\ 
  \textit{sehr} & \textit{sehr} & I am {very} \underline{easy} and {very} modest.   \\ 
\\
 
&& \textcolor{mygray}{Reference}\\
\hline
 $\phi$ & $\phi$   & \textit{I am relieved and humble.}\\
  \textit{sehr} & \textit{sehr} & \textit{I am very relieved and very humble.}   \\ 
\end{tabularx}
\caption{Insertion of the German word \textit{sehr} (English: \textit{very}) in different positions in the source sentence results in substantially different translations. 
$^\ddagger$ indicates the original sentence from WMT 2017.}
\label{first}
\end{table}

Recently, several approaches investigated NMT models when encountering noisy input and how \textit{worst-case examples} of noisy input can `break' state-of-the-art NMT models \citep{goodfellow6572explaining,D18-1050}. 
\citet{DBLP:journals/corr/abs-1711-02173} show that character-level noise in the input 
leads to poor translation performance.
\citet{halluc} randomly insert words in different positions in the source sentence and observe that in some cases the translations are completely unrelated to the input. 
While it is to some extent expected that the performance of NMT models that are trained on predominantly clean but tested on noisy data deteriorates, other changes are more unexpected.

In this paper, we explore unexpected and erroneous changes in the output of NMT models.
Consider the simple example in Table~\ref{first} where the Transformer model \citep{vaswani2017attention}
is used to translate very similar sentences.
Surprisingly, we observe that by simply altering one word in the source sentence---inserting the German word \textit{sehr} (English: \textit{very})---an unrelated change occurs in the translation.
In principle, an NMT model that generates the translation of the word \textit{erleichtert} (English: \textit{relieved}) in one context, should also be able to generalize and translate it correctly in a very similar context.
Note that there are no infrequent words in the source sentence and after each modification, the input is still syntactically correct and semantically plausible.
We call a model \textit{volatile} if it displays inconsistent behaviour across similar input sentences during inference. 

\begin{table*}[htb!]
\begin{center}\small
\setlength\tabcolsep{6pt} 
\begin{tabularx}{\textwidth}{lX}
\toprule
Modification & Sentence variations\\
\midrule
\texttt{DEL} &  Some 500 years after the Reformation, Rome  \textbf{[now\textbackslash$\pmb{\phi}$]} has a Martin Luther Square. \\
\texttt{SUBNUM} & I'm very pleased for it to have happened at Newmarket because this is where I landed \textbf{[30\textbackslash31]} years ago. \\
\texttt{INS} & I loved Amy and she is  \textbf{[$\pmb{\phi}$\textbackslash also]} the only person who ever loved me.  \\
\texttt{SUBGEN} & \textbf{[He\textbackslash She]} received considerable appreciation and praise for this. \\
\bottomrule
\end{tabularx}
\end{center}
\caption{\label{examplesofpert} Examples of different variations from WMT. [$w_i$\textbackslash$w_j$] indicates that $w_i$ in the original sentence is replaced by $w_j$. ${\phi}$ is an empty string.}
\end{table*} 

We investigate to what extent well-established NMT models are volatile during inference.
Specifically, we locally modify sentence pairs in the test set and identify examples where a trivial modification in the source sentence causes an `unexpected change' in the translation.
These modifications are generated conservatively to avoid insertion of any noise or rare words in the data (Section~\ref{classs}). 
Our goal is not to \textit{fool} NMT models, but instead identify common cases where the models exhibit unexpected behaviour and in the worst cases result in incorrect translations.

We observe that our modifications expose volatilities of both RNN and Transformer translation models in $26\%$ and $19\%$ of sentence variations, respectively.
Our findings show how vulnerable current NMT models are to trivial linguistic variations, putting into question the  generalizability of these models.

\section{Sentence Variations}

\subsection{Is this another noisy text translation problem?}

Noisy input text can cause mistranslations in most MT systems, and there has been growing research interest in studying the behaviour of MT systems when encountering noisy input \citep{li-EtAl:2019:WMT1}.

\citet{DBLP:journals/corr/abs-1711-02173} propose to  swap or randomize letters in a word in the input sentence. For instance, they change the word \textit{noise} in the source sentence into \textit{iones}.
\citet{halluc} examine how the insertion of a random word in a random position in the source sentence leads to mistranslations. 
\citet{D18-1050} proposes a benchmark dataset for translation of noisy input sentences, consisting of noisy, user-generated comments on Reddit.
The types of noisy input text they observe include spelling or typographical errors, word omission/insertion/repetition, and grammatical errors.

In these previous works, the focus of the research is on studying how the MT systems are not robust when handling noisy input text. In these approaches, the input sentences are semantically or syntactically incorrect which leads to mistranslations. 

However, in this paper, our focus is on input text that does \textit{not} contain any types of noise. We modify input sentences in a way that the outcomes are still syntactically and semantically correct. 
We investigate how the MT systems exhibit volatile behaviour in translating sentences that are extremely similar and only differ in one word without any noise injection.

\subsection{Variation generation} \label{classs}

While there are various ways to automatically modify sentences, we are interested in simple semantic and syntactic modifications. 
These trivial linguistic variations should have almost no effect on the translation of the rest of the sentence.

 \begin{figure*}[htb!]
\centering
\includegraphics[width=\textwidth]{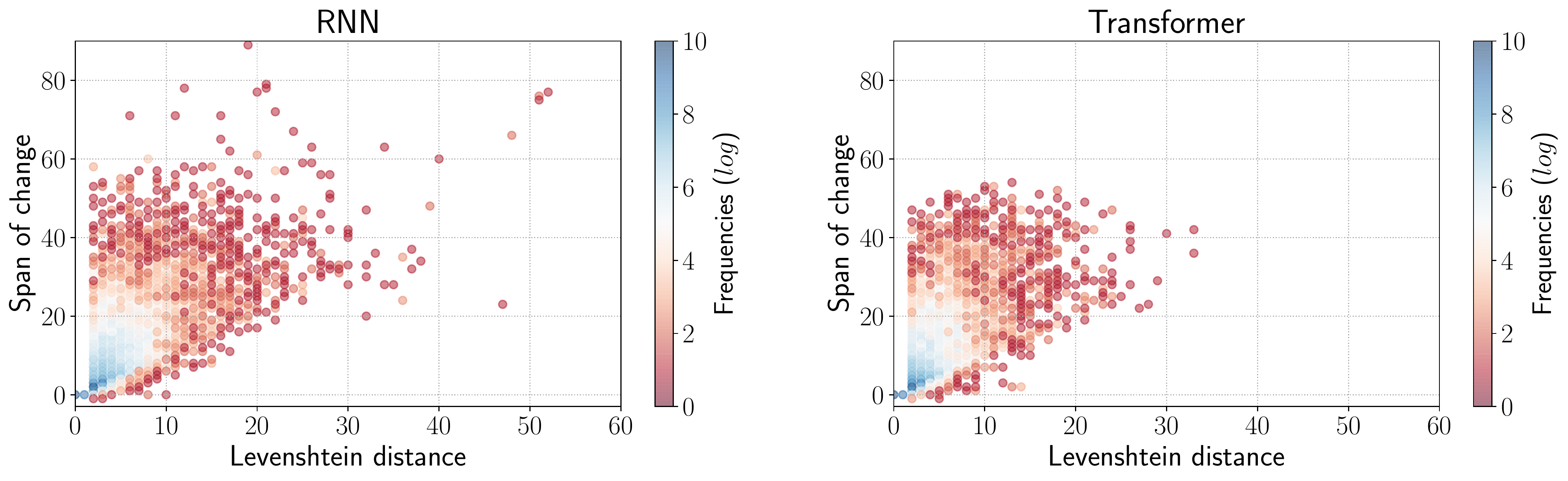}
\caption{\textit{Levenshtein distance} and \textit{span of change} between translations of sentence variations for RNN and Transformer. 
The majority of sentence variations falls into the category of \textit{minor} changes between translations (blue area). However, a surprising number of cases have significant changes (red area). RNN exhibits a slightly more unstable pattern i.e., sentence variations with large edit differences and large spans of change.
}
\label{fig1}
\end{figure*}


We define a set of rules to slightly modify the source and target sentences in the test data and keep the sentences syntactically correct and semantically plausible. 


\paragraph{\textnormal{ \texttt{DEL}}} A conservative approach of modifying a sentence automatically without breaking the grammaticality of a sentence is to remove adverbs. 
We identify a list of the 50 most frequent adverbs in English and their translations in German\footnote{\url{dict.cc}}.
For every sentence in the WMT test sets, if we find a sentence pair containing both a word and its translation from this list, we remove both words and create a new sentence pair. 

\paragraph{\textnormal{ \texttt{SUBNUM}}} Another simple yet effective approach to safely modify sentences 
is to substitute numbers with other numbers.
In this approach, we select every sentence pair from the test sets that contains a number and substitute the number $i$ in both source and target sentences with $i+k$ where $1 \leq k \leq 5$.
We choose a small range for change so that the sentences are still semantically correct for the most part and result in few implausible sentences.

\paragraph{\textnormal{ \texttt{INS}}} Randomly inserting words in a sentence has a high chance of producing a syntactically incorrect sentence.
To ensure that sentences remain grammatical and semantically plausible after modification, we define a bidirectional $n$-gram probability for inserting new words as follows:
\begin{equation*}
P(w_3 | w_1 w_2 w_4 w_5) = \frac{C(w_1 w_2 w_3 w_4 w_5)}{C(w_1 w_2 \sbullet[1] w_4 w_5)}
\end{equation*}
$w_3$ is inserted in the middle of the phrase $w_1 w_2 w_4 w_5$, if the conditional probability is greater than a predefined threshold.  
The probabilities are computed on the WMT data.
This simple approach, instead of using a more complex language model, serves our purposes since we are interested in inserting very common words that are already captured by the $n$-grams in the training data.

\paragraph{ \textnormal{\texttt{SUBGEN}}} Finally, a local modification 
is changing the gender of the person in the sentences. 
The goal of this modification is to investigate the existence and severity of gender bias in our models.
This is inspired by recent approaches that have shown that NMT models learn social stereotypes such as gender bias from training data \citep{escude-font-costa-jussa-2019-equalizing,stanovsky-etal-2019-evaluating}.  
\\
\\
Note that in a minority of cases these procedures can lead to semantically incorrect sentences, for instance, by substituting numbers we can potentially generate sentences such as \textit{``She was born on October 34th``}. 
While this can cause problems for a reasoning task, it barely affects the translation task, as long as the modifications are consistent on the source and target side. 

Table~\ref{examplesofpert} shows examples of generated variations. 
We emphasize that only modifications with local consequences have been selected and we intentionally ignore cases such as \textit{negation} which can result in wider structural changes in the translation of the sentence.

\begin{table}[htb!]
\center
\small
\setlength\tabcolsep{4.5pt} 
\begin{tabularx}{\columnwidth}{lccc|ccc}
\toprule
 & \multicolumn{3}{c}{\bf DE-EN} &  \multicolumn{3}{|c}{\bf EN-DE} \\
\midrule   &  2016 & 2017 & 2018  & 2016 & 2017  & 2018  \\ \midrule
RNN & 32.5 & 28.2 &35.2 & 28.1  & 22.4 & 34.6 \\ 
Transformer & 36.2 & 32.1 & 40.1 & 33.4 & 27.9& 39.8  \\
\bottomrule
\end{tabularx}
\caption{\label{bleus} BLEU scores for different models on the WMT data for translation DE$\leftrightarrow$EN.}
\end{table}

\begin{table*}[htb!]
 \begin{small}
\begin{center}
\setlength\tabcolsep{4pt} 
\caption{\label{man}  A random sample of sentences from the WMT test sets and our proposed variations shown with `unexpected change' annotations ($\Delta Translation$). The cases where the unexpected change leads to a change in translation quality are marked in column $\Delta Quality$. \textcolor{colorzero}{\textbf{[$w_i$\textbackslash$w_j$]}} indicates that $w_i$ in the original sentence is replaced by $w_j$. $S$ is the original and modified source sentence, $R$ is the original and modified reference translation, $T$ is the translation of the original sentence, and $T_m$ is the translation of the modified sentence. Differences in translations related to annotations in the original and the modified translations are in \textcolor{colorone}{\textbf{red}} and \textcolor{colortwo}{\textbf{orange}}, respectively. Note that we are interested in \textit{unexpected} changes and do not highlight the changes that are a direct consequence of the modifications.}
\begin{tabularx}{\textwidth}{lX}
\toprule
${S}$ &   Coes letztes Buch ``Chop Suey" handelte von der chinesischen K{\"u}che in den USA, w{\"a}hrend Ziegelman in ihrem Buch ``\textcolor{colorzero}{\textbf{{{[97\textbackslash 101]}}}} Orchard" {\"u}ber das Leben in einem Wohnhaus an der Lower East Side aus der Lebensmittelperspektive erz{\"a}hlt.  \\
${R}$ & Mr. Coe's last book, ``Chop Suey," was about Chinese cuisine in America, while Ms. Ziegelman told the story of life in a Lower East Side tenement through food in her book ``\textcolor{colorzero}{\textbf{{[97\textbackslash 101]}}} Orchard."  \\ 
${T}$ &   Coes\textcolor{colorone}{\textbf{{'s}}} last book, ``Chop Suey\textcolor{colorone}{\textbf{{,}}}" was about Chinese cuisine in the \textcolor{colorone}{\textbf{{US}}}, while Ziegelman, \textcolor{colorone}{\textbf{{in her book ``97 Orchard" talks about}}} living in a lower East Side.  \\
${T_m}$ &  Coes last book ``Chop Suey" was about Chinese cuisine in the \textcolor{colortwo}{\textbf{{United States}}}, while Ziegelman \textcolor{colortwo}{\textbf{{writes in her book ``101 Orchard" about}}} living in a lower East Side.\vspace{.3cm} \\ 
  \rowcolor{tablegray}   \multicolumn {2}{l}{\textcolor{mygray}{\texttt{$\Delta Translation$: [reordered] [paraphrased]  }}}  \\
      \rowcolor{tablegray}   \multicolumn {2}{l}{\textcolor{mygray}{\texttt{$\Delta Quality$: No }}}  \\
  \midrule
 ${S}$ &  Man h{\"a}lt \textcolor{colorzero}{\textbf{{[bereits\textbackslash$\pmb{\phi}$]}}} Ausschau nach Parkbank, Hund und Fu{\ss}ball spielenden Jungs und M{\"a}dels. \\
$R$ & You are \textcolor{colorzero}{\textbf{{[already\textbackslash$\pmb{\phi}$]}}} on the lookout for a park bench, a dog, and boys and girls playing football. \\
${T}$ & \textcolor{colorone}{\textbf{We are}} already \textcolor{colorone}{\textbf{looking}} for Parkbank, dog and football playing boys and girls.  \\
${T_m}$& \textcolor{colortwo}{\textbf{Look}} for Parkbank, dog and football playing boys and girls. \vspace{.3cm} \\ 
  \rowcolor{tablegray}   \multicolumn {2}{l}{\textcolor{mygray}{\texttt{$\Delta Translation$: [word form] [add/remove]}}}  \\
      \rowcolor{tablegray}   \multicolumn {2}{l}{\textcolor{mygray}{\texttt{$\Delta Quality$: Yes }}}  \\
  \midrule
${S}$ 	& Bei einem Unfall eines Reisebusses mit \textcolor{colorzero}{\textbf{{[43\textbackslash 45]}}} Senioren als Fahrg{\"a}sten sind am Donnerstag in Krummh{\"o}rn (Landkreis Aurich) acht Menschen verletzt worden.  \\
$R$ & On Thursday, an accident involving a coach carrying \textcolor{colorzero}{\textbf{{[43\textbackslash 45]}}} elderly people in Krummh{\"o}rn (district of Aurich) led to eight people being injured. \\
${T}$  &	In the event of an accident involving \textcolor{colorone}{\textbf{a coach with 43 senior citizens}} as \textcolor{colorone}{\textbf{passengers}}, eight people were injured on Thursday in \textcolor{colorone}{\textbf{Krummaudin (County Aurich)}}.  \\
${T_m}$   &	In the event of an accident involving \textcolor{colortwo}{\textbf{a 45-year-old coach}} as a \textcolor{colortwo}{\textbf{passenger}}, eight people were injured on Thursday in  \textcolor{colortwo}{\textbf{the district of Aurich}}. \vspace{.3cm} \\
  \rowcolor{tablegray} \multicolumn {2}{l}{\textcolor{mygray}{\texttt{$\Delta Translation$: [word form] [add/remove] [other] }}}  \\
      \rowcolor{tablegray}   \multicolumn {2}{l}{\textcolor{mygray}{\texttt{$\Delta Quality$: Yes }}}  \\
\midrule
${S}$  &  Es ist ein anstrengendes Pensum, aber die Dorfmusiker helfen \textcolor{colorzero}{\textbf{{[normalerweise\textbackslash$\pmb{\phi}$]}}}, das Team motiviert zu halten.  \\	
 $R$ & It's a backbreaking pace, but village musicians \textcolor{colorzero}{\textbf{{[usually\textbackslash$\pmb{\phi}$]}}} help keep the team motivated.    \\
${T}$  & It\textcolor{colorone}{\textbf{'s}} a \textcolor{colorone}{\textbf{demanding child}}, but the village \textcolor{colorone}{\textbf{musicians}} usually \textcolor{colorone}{\textbf{help}} keep the team motivated.  \\
${T_m}$  & It \textcolor{colortwo}{\textbf{is}} a \textcolor{colortwo}{\textbf{hard-to-use}}, but the village \textcolor{colortwo}{\textbf{musician helps to}} keep the team motivated.\vspace{.3cm}	\\ 
  \rowcolor{tablegray}   \multicolumn {2}{l}{\textcolor{mygray}{\texttt{$\Delta Translation$: [word form] [other]}}}  \\
    \rowcolor{tablegray}   \multicolumn {2}{l}{\textcolor{mygray}{\texttt{$\Delta Quality$: Yes }}}  \\
\bottomrule
\end{tabularx}
 \end{center}
 \end{small}
\end{table*}

We generate $10k$ sentence variations by applying these modifications to all sentence pairs in WMT test sets 2013--2018 \citep{bojar-EtAl:2018:WMT1}. 
We use RNN and Transformer models to translate sentences and their variations.

\subsection{Experimental setup}

In the translation experiments, we use the standard EN$\leftrightarrow$DE WMT-2017 training data 
\citep{bojar-EtAl:2018:WMT1}.
We perform NMT experiments with two different architectures: RNN \citep{luong:2015:EMNLP} and Transformer \citep{vaswani2017attention}.
We preprocess the training data with Byte-Pair Encoding (BPE) using 32K merge operations \citep{sennrich-haddow-birch:2016:P16-12}. 
During inference, we use beam search with a beam size of 5.
Table~\ref{bleus} shows the case-sensitive BLEU scores as calculated by \texttt{multi-bleu.perl}. 

\paragraph{RNN} As the first NMT system, we use a 2-layer bidirectional attention-based LSTM model implemented in OpenNMT \citep{2017opennmt} trained with an embedding size of 512, hidden dimension size of 1024, and batch size of 64 sentences. We use Adam \citep{kingma2014adam} for optimization.

\paragraph{Transformer} We also experiment with the Transformer model \citep{vaswani2017attention} implemented in OpenNMT. We train a model with 6 layers, the hidden size is set to 512 and the filter size is set to 2048. The multi-head attention has 8  attention heads. We use Adam \citep{kingma2014adam} for optimization. All parameters are set based on the suggestions in \citet{2017opennmt} to replicate the results of the original paper.

\begin{figure*}[htb!]
\centering
\includegraphics[width=1\textwidth]{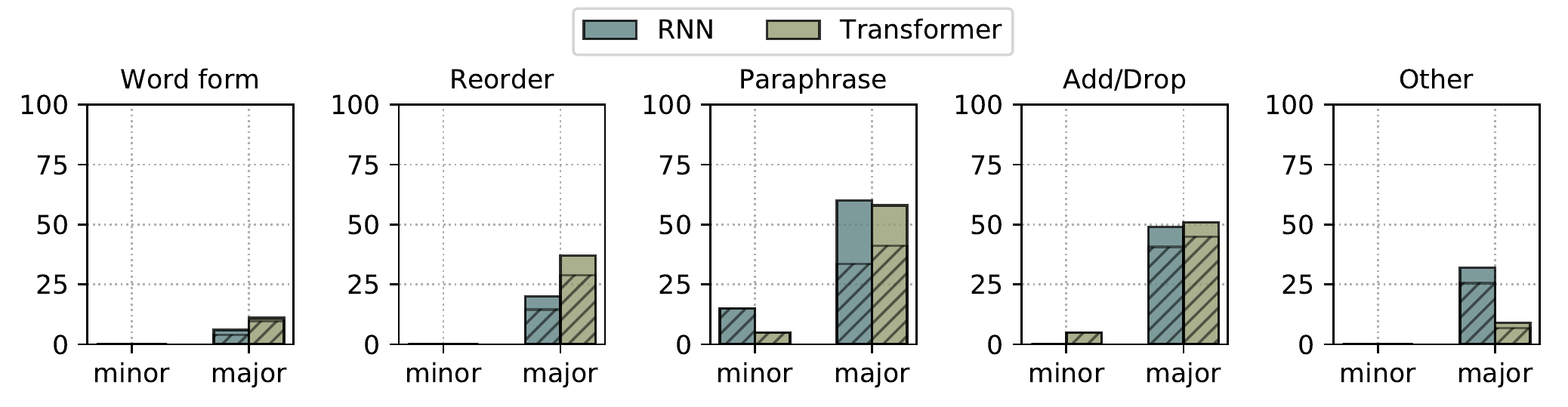}
\caption{Categories of \textit{unexpected changes} in the translation of sentence variations as provided by annotators. The percentage of sentence variations with \textit{minor} and \textit{major} edit differences, as defined in~\ref{ctsec}, are shown separately. The hatched pattern indicates the ratio of sentence variations for which the translation quality changes.
Note that \textit{expected changes} are not plotted here.
} 
\label{fig2}
\end{figure*}

\section{Evaluation of unexpected and erroneous changes} 

The modifications described above generate sentences that are extremely similar 
and hence are expected to have a very similar difficulty of translation. 
We evaluate the NMT models on how robust and consistent they are in translating these sentence variations rather than their absolute quality.

\subsection{Deviations from Original Translations} \label{ctsec}

The variations are aimed to have minimal effect on changing the meaning of the sentences. 
Hence, major changes in the translations of these variations can be an indication of volatility in the model.
To assess whether the proposed sentence variations result in major changes in the translations, we measure changes in the translations of sentence variations with Levenshtein distance \citep{levenshtein1966binary}. 
Specifically, Levenshtein distance measures the edit distance between the two translations. 
We also use the first and last positions of change in the translations, which represents the span of changes.

Ideally, with our simple modifications, we expect a value of zero for the span of change and a value of at most 2 for the Levenshtein distance for a translation pair.
This indicates that there is only one token difference between the translation of the original sentence and the modified sentence. 
We define two types of changes based on these measures: \textit{minor} and \textit{major}.
We choose the threshold to distinguish between minor and major changes more conservatively to allow for more variations in the translations.
The change in translations is empirically considered \textit{major} if both metrics are greater than 10, and \textit{minor} if both are less than 10. 
Note that edit distances and spans are based on BPE subword units.

With two very similar source sentences, we expect the Levenshtein distance and span of change between translations of these sentences to be small.
Figure~\ref{fig1} shows the results for the RNN and Transformer model. 
While the majority of sentence variations have minor changes, 
a substantial number of sentences, $18\%$ of RNN and $13\%$ of Transformer translations, result in translations with major differences. 
This is surprising and an indication of volatility since these trivial modifications, in principle, should only result in minor and local changes in the translations.

\subsection{Oscillations of Variation in Translations}

In this section, we look into various sentence-level metrics to further analyze the observed behaviour. 
In particular, we focus on the \texttt{SUBNUM} modification because with this modification we can generate numerous variations of the same sentence. 
Having a high number of variations for each sentence gives us the opportunity of observing oscillations of various string matching metrics. 

We use sentence-level BLEU, METEOR \citep{denkowski-lavie-2011-meteor}, TER \citep{Snover06astudy}, and LengthRatio to quantify changes in the translations.
LengthRatio represents the translation length over reference length as a percentage.
For a given source sentence, we define the oscillation range as changes in the sentence-level metric for
 the translations of variations of a given sentence. 

\begin{figure*}[htb!]
\centering
\includegraphics[width=1\textwidth]{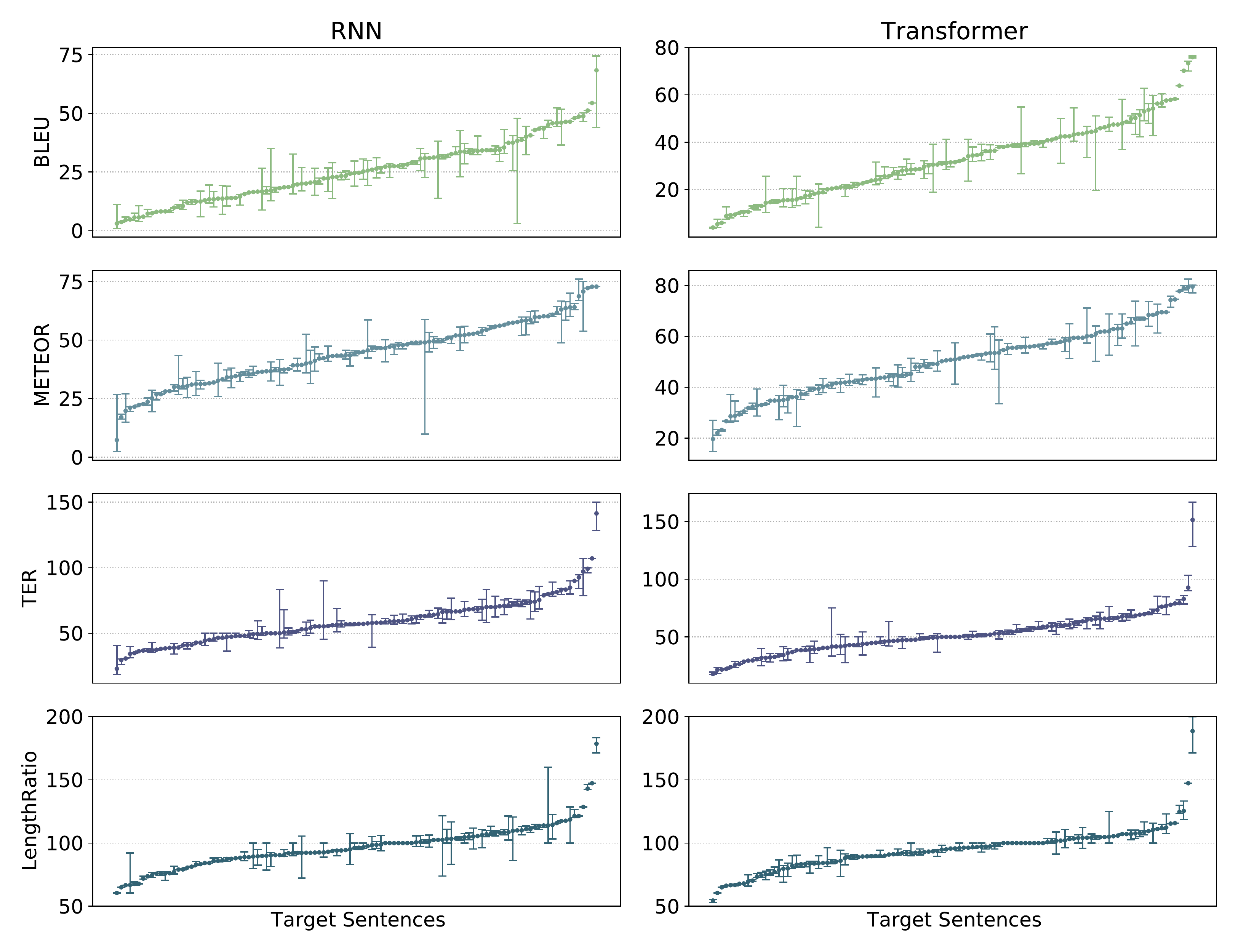}
\caption{Oscillations of various sentence-level attributes for randomly sampled sentences from our test data and their \texttt{SUBNUM} variations. 
The data points are the mean values for all variations of each sentence, and the error bars indicate the range of oscillation of the metrics. The x-axis represents test sentence instances, sorted based on the corresponding metric. Ideally each data point should have zero oscillation. }
\label{figosc}
\end{figure*}

While sentence-level metrics are not reliable indicators of translation quality, they do capture fluctuations in translations.  
With the variations we introduce, in theory there should be no fluctuations in the translations. 
Table~\ref{oscstats} and Figure~\ref{figosc} provide the results. 
We observe that even though these sentence variations differ by only one number, there are many cases where an insignificant change in the sentence results in unexpectedly large oscillations. Both RNN and Transformer exhibit this behaviour to a certain extent. 

\begin{table}[ht]
\center
\small
\setlength{\tabcolsep}{5pt}
\caption{\label{oscstats} Mean oscillations for \texttt{SUBNUM} variations. In theory the variations should result in zero oscillations for every metric.}
\begin{tabularx}{\columnwidth}{lcccc}
\toprule
& BLEU & METEOR & TER & LengthRatio \\
\midrule
RNN & 4.0 & 3.8& 5.2 & 5.3 \\ 
Transformer & 3.8 & 3.3 & 4.2 & 3.4  \\
\bottomrule
\end{tabularx}
\end{table}

\subsection{The Effect of Volatility on Translation Quality}

While edit distances and spans of change provide some indication of volatility, they do not capture all aspects of this unexpected behaviour. 
It is also not entirely clear what effect these unexpected changes have 
on translation quality.
To further investigate 
this, we also perform manual evaluations. 
 
In the first evaluation, we provide annotators with a pair of sentence variations and their corresponding translations
and ask them to identify the differences between the two sentence pairs.
In the second evaluation, we additionally provide the source sentences and reference translations, and ask the annotators to rank the sentence variations based on the translation quality similar to \citet{bojar-EtAl:2016:WMT1}.
In total the annotators evaluated 400 randomly selected sentence quadruplets. 

The annotators identified $71\%$ and $68\%$ of changes in the variation translation as \textit{expected} for the RNN and Transformer model, respectively.
The main types of \textit{unexpected} changes identified by the annotators are a change of word form, e.g., verb tense,, reordering of phrases, paraphrasing parts of the sentence, and an `other' category, e.g., preposition.  
A sentence pair can have multiple labels based on the types of changes. 
Table~\ref{man} provides examples from the test data. 

Statistics for each category of unexpected change is shown in Figure~\ref{fig2}.
Our first observation is that, as to be expected, there are very few `unexpected changes' when two variations lead to translations with \textit{minor} differences.
Interestingly, the vast majority of changes are due to paraphrasing and dropping of words.
Comparing the performance of the RNN and Transformer model, we see that both RNN and Transformer display inconsistent translation behaviour. 
While Transformer has slightly fewer sentences with major changes, it has a higher number of sentence variations in the major category that result in a change in translation quality. 
From the annotators' assessments, we find that in $26\%$ and $19\%$ of sentence variations, the modification results in a change in \textit{translation quality} for the RNN and Transformer model, respectively.




 
\subsection{Generalization and Compositionality}

Because of their ability to generalize beyond their training data, deep learning models achieve exceptional performances in numerous tasks.
The generalization ability allows MT systems to generate long sentences not seen before. 
Recently there has been some interest in understanding whether this performance depends on recognizing
shallow patterns, or whether the networks are indeed capturing and generalizing linguistic rules.


In simple terms, compositionality is the ability to construct larger linguistic expressions by combining simpler parts. For instance, if a model understands the correct compositional rules to understand \textit{`John loves Mary'}, it must also understand \textit{`Mary loves John'} \citep{fodor2002compositionality}.
Investigating the compositional behaviour of neural networks in real-world natural language problems is a challenging task. 
Recently, several works have studied deep learning models' understanding of compositionality in natural language by using synthetic and simplified languages \citep{DBLP:journals/corr/abs-1902-07181,babyai_iclr19}.
\citet{DBLP:journals/corr/abs-1904-00157} shows that to a certain extent neural networks can be productive without being compositional.

Although we do not specifically look into the compositional potential of MT systems, we are inspired by compositionality in defining our modifications.
We argue that the observed volatile behaviour of the MT systems in this paper is a side effect of current models not being compositional.  
If an MT system has a good `understanding' of the underlying structures of the sentences \textit{`Mary is 10 years old'} and \textit{`Mary is 11 years old'}, it must also translate them very similarly regardless of the accuracy of the translation. 
While current evaluation metrics capture the accuracy of the NMT models, these volatilities go unnoticed.  

Current neural models are successful in generalizing without learning any explicit compositional rules, however, our findings signal that they still lack robustness.
We highlight this lack of robustness and suspect that it is associated with these models' lack of understanding of the compositional nature of language.

\section{Conclusion}

In this paper, we showed the unexpected volatility of NMT models by using a simple approach to modifying standard test sentences without introducing noise, i.e., by generating semantically and syntactically correct variations.
We show that even with trivial linguistic modifications of source sentences we can effectively identify a surprising number of cases where the translations of extremely similar sentences are surprisingly different, see Figure~\ref{fig1}.
Our manual analyses show that both RNN and Transformer models exhibit volatile behaviour with changes in translation quality for $26\%$ and $19\%$ of sentence variations, respectively.
This highlights the problem of generalizability of current 
NMT models and we hope that our insights will be useful for developing more robust NMT models.

\section*{Acknowledgments}

We thank Arianna Bisazza for helpful discussions.
This research was funded in part by the
Netherlands Organization for Scientific Research
(NWO) under project numbers 639.022.213 and
612.001.218. We also thank NVIDIA for their
hardware support and the anonymous reviewers
for their helpful comments.

\bibliography{acl2018}

\begin{thebibliography}{29}
\expandafter\ifx\csname natexlab\endcsname\relax\def\natexlab#1{#1}\fi

\bibitem[{Andreas(2019)}]{DBLP:journals/corr/abs-1902-07181}
Jacob Andreas. 2019.
\newblock \href {http://arxiv.org/abs/1902.07181} {Measuring compositionality
  in representation learning}.
\newblock \emph{CoRR}, abs/1902.07181.

\bibitem[{Baroni(2019)}]{DBLP:journals/corr/abs-1904-00157}
Marco Baroni. 2019.
\newblock \href {http://arxiv.org/abs/1904.00157} {Linguistic generalization
  and compositionality in modern artificial neural networks}.
\newblock \emph{CoRR}, abs/1904.00157.

\bibitem[{Belinkov and Bisk(2018)}]{DBLP:journals/corr/abs-1711-02173}
Yonatan Belinkov and Yonatan Bisk. 2018.
\newblock \href {http://people.csail.mit.edu/belinkov/assets/pdf/iclr2018.pdf}
  {Synthetic and natural noise both break neural machine translation}.
\newblock In \emph{Proceedings of the International Conference on Learning
  Representations (ICLR)}.

\bibitem[{Bojar et~al.(2016)Bojar, Chatterjee, Federmann, Graham, Haddow, Huck,
  Jimeno~Yepes, Koehn, Logacheva, Monz, Negri, Neveol, Neves, Popel, Post,
  Rubino, Scarton, Specia, Turchi, Verspoor, and
  Zampieri}]{bojar-EtAl:2016:WMT1}
Ond\v{r}ej Bojar, Rajen Chatterjee, Christian Federmann, Yvette Graham, Barry
  Haddow, Matthias Huck, Antonio Jimeno~Yepes, Philipp Koehn, Varvara
  Logacheva, Christof Monz, Matteo Negri, Aurelie Neveol, Mariana Neves, Martin
  Popel, Matt Post, Raphael Rubino, Carolina Scarton, Lucia Specia, Marco
  Turchi, Karin Verspoor, and Marcos Zampieri. 2016.
\newblock \href {http://www.aclweb.org/anthology/W/W16/W16-2301} {Findings of
  the 2016 conference on machine translation}.
\newblock In \emph{Proceedings of the First Conference on Machine Translation},
  pages 131--198, Berlin, Germany. Association for Computational Linguistics.

\bibitem[{Bojar et~al.(2018)Bojar, Federmann, Fishel, Graham, Haddow, Huck,
  Koehn, and Monz}]{bojar-EtAl:2018:WMT1}
Ond\v{r}ej Bojar, Christian Federmann, Mark Fishel, Yvette Graham, Barry
  Haddow, Matthias Huck, Philipp Koehn, and Christof Monz. 2018.
\newblock \href {http://www.aclweb.org/anthology/W18-6401} {Findings of the
  2018 conference on machine translation (wmt18)}.
\newblock In \emph{Proceedings of the Third Conference on Machine Translation,
  Volume 2: Shared Task Papers}, pages 272--307, Belgium, Brussels. Association
  for Computational Linguistics.

\bibitem[{Chevalier-Boisvert et~al.(2019)Chevalier-Boisvert, Bahdanau, Lahlou,
  Willems, Saharia, Nguyen, and Bengio}]{babyai_iclr19}
Maxime Chevalier-Boisvert, Dzmitry Bahdanau, Salem Lahlou, Lucas Willems,
  Chitwan Saharia, Thien~Huu Nguyen, and Yoshua Bengio. 2019.
\newblock \href {https://openreview.net/forum?id=rJeXCo0cYX} {Baby{AI}: First
  steps towards grounded language learning with a human in the loop}.
\newblock In \emph{International Conference on Learning Representations}.

\bibitem[{Denkowski and Lavie(2011)}]{denkowski-lavie-2011-meteor}
Michael Denkowski and Alon Lavie. 2011.
\newblock \href {https://www.aclweb.org/anthology/W11-2107} {Meteor 1.3:
  Automatic metric for reliable optimization and evaluation of machine
  translation systems}.
\newblock In \emph{Proceedings of the Sixth Workshop on Statistical Machine
  Translation}, pages 85--91, Edinburgh, Scotland. Association for
  Computational Linguistics.

\bibitem[{Escud{\'e}~Font and
  Costa-juss{\`a}(2019)}]{escude-font-costa-jussa-2019-equalizing}
Joel Escud{\'e}~Font and Marta~R. Costa-juss{\`a}. 2019.
\newblock \href {https://doi.org/10.18653/v1/W19-3821} {Equalizing gender bias
  in neural machine translation with word embeddings techniques}.
\newblock In \emph{Proceedings of the First Workshop on Gender Bias in Natural
  Language Processing}, pages 147--154, Florence, Italy. Association for
  Computational Linguistics.

\bibitem[{Fadaee et~al.(2018)Fadaee, Bisazza, and Monz}]{L18-1148}
Marzieh Fadaee, Arianna Bisazza, and Christof Monz. 2018.
\newblock \href {http://aclweb.org/anthology/L18-1148} {Examining the tip of
  the iceberg: A data set for idiom translation}.
\newblock In \emph{Proceedings of the Eleventh International Conference on
  Language Resources and Evaluation (LREC-2018)}. European Language Resource
  Association.

\bibitem[{Fodor and LePore(2002)}]{fodor2002compositionality}
J.A. Fodor and E.~LePore. 2002.
\newblock \href {https://books.google.nl/books?id=qcI6IVquMdgC} {\emph{The
  Compositionality Papers}}.
\newblock Clarendon Press.

\bibitem[{Frege(1892)}]{Frege1892}
G.~Frege. 1892.
\newblock "uber sinn und bedeutung.
\newblock In Mark Textor, editor, \emph{Funktion - Begriff - Bedeutung},
  volume~4 of \emph{Sammlung Philosophie}. Vandenhoeck \& Ruprecht, G"ottingen.

\bibitem[{Goodfellow et~al.(2015)Goodfellow, Shlens, and
  Szegedy}]{goodfellow6572explaining}
Ian~J Goodfellow, Jonathon Shlens, and Christian Szegedy. 2015.
\newblock \href {https://arxiv.org/pdf/1412.6572.pdf} {Explaining and
  harnessing adversarial examples}.
\newblock In \emph{Proceedings of the International Conference on Learning
  Representations (ICLR)}.

\bibitem[{Janssen(2001)}]{DBLP:journals/jolli/Janssen01}
Theo M.~V. Janssen. 2001.
\newblock \href {https://doi.org/10.1023/A:1026542332224} {Frege, contextuality
  and compositionality}.
\newblock \emph{Journal of Logic, Language and Information}, 10(1):115--136.

\bibitem[{Khayrallah and Koehn(2018)}]{W18-2709}
Huda Khayrallah and Philipp Koehn. 2018.
\newblock \href {http://aclweb.org/anthology/W18-2709} {On the impact of
  various types of noise on neural machine translation}.
\newblock In \emph{Proceedings of the 2nd Workshop on Neural Machine
  Translation and Generation}, pages 74--83. Association for Computational
  Linguistics.

\bibitem[{Kingma and Ba(2015)}]{kingma2014adam}
Diederik~P Kingma and Jimmy~Lei Ba. 2015.
\newblock \href {https://arxiv.org/pdf/1412.6980.pdf} {Adam: A method for
  stochastic optimization}.
\newblock In \emph{Proceedings of the International Conference on Learning
  Representations (ICLR)}.

\bibitem[{Klein et~al.(2017)Klein, Kim, Deng, Senellart, and
  Rush}]{2017opennmt}
Guillaume Klein, Yoon Kim, Yuntian Deng, Jean Senellart, and Alexander Rush.
  2017.
\newblock \href {http://www.aclweb.org/anthology/P17-4012} {Opennmt:
  Open-source toolkit for neural machine translation}.
\newblock In \emph{Proceedings of ACL 2017, System Demonstrations}, pages
  67--72. Association for Computational Linguistics.

\bibitem[{Koehn and Knowles(2017)}]{koehn2017six}
Philipp Koehn and Rebecca Knowles. 2017.
\newblock Six challenges for neural machine translation.
\newblock \emph{arXiv preprint arXiv:1706.03872}.

\bibitem[{Lee et~al.(2018)Lee, Firat, Agarwal, Fannjiang, and
  Sussillo}]{halluc}
Katherine Lee, Orhan Firat, Ashish Agarwal, Clara Fannjiang, and David
  Sussillo. 2018.
\newblock \href {https://openreview.net/pdf?id=SkxJ-309FQ} {Hallucinations in
  neural machine translation}.
\newblock In \emph{Neural Information Processing Systems (NeurIPS) Workshop on
  Interpretability and Robustness for Audio, Speech, and Language}. NeurIPS.

\bibitem[{Levenshtein(1966)}]{levenshtein1966binary}
Vladimir~I Levenshtein. 1966.
\newblock \href {https://nymity.ch/sybilhunting/pdf/Levenshtein1966a.pdf}
  {Binary codes capable of correcting deletions, insertions, and reversals}.
\newblock \emph{Soviet physics doklady}, 10(8):707--710.

\bibitem[{Li et~al.(2019)Li, Michel, Anastasopoulos, Belinkov, Durrani, Firat,
  Koehn, Neubig, Pino, and Sajjad}]{li-EtAl:2019:WMT1}
Xian Li, Paul Michel, Antonios Anastasopoulos, Yonatan Belinkov, Nadir Durrani,
  Orhan Firat, Philipp Koehn, Graham Neubig, Juan Pino, and Hassan Sajjad.
  2019.
\newblock Findings of the first shared task on machine translation robustness.
\newblock In \emph{Proceedings of the Fourth Conference on Machine Translation
  (Volume 2: Shared Task Papers, Day 1)}, pages 91--102, Florence, Italy.
  Association for Computational Linguistics.

\bibitem[{Luong et~al.(2015)Luong, Pham, and Manning}]{luong:2015:EMNLP}
Thang Luong, Hieu Pham, and Christopher~D. Manning. 2015.
\newblock \href {http://aclweb.org/anthology/D15-1166} {Effective approaches to
  attention-based neural machine translation}.
\newblock In \emph{Proceedings of the 2015 Conference on Empirical Methods in
  Natural Language Processing}, pages 1412--1421, Lisbon, Portugal. Association
  for Computational Linguistics.

\bibitem[{Michel and Neubig(2018)}]{D18-1050}
Paul Michel and Graham Neubig. 2018.
\newblock \href {http://aclweb.org/anthology/D18-1050} {Mtnt: A testbed for
  machine translation of noisy text}.
\newblock In \emph{Proceedings of the 2018 Conference on Empirical Methods in
  Natural Language Processing}, pages 543--553. Association for Computational
  Linguistics.

\bibitem[{Montague(1974)}]{montague1974d}
Richard Montague. 1974.
\newblock Universal grammar.
\newblock In \emph{Formal Philosophy: Selected Papers of Richard Montague},
  number 222--247 in Theoria, New Haven, London. Yale University Press.

\bibitem[{Pelletier(1994)}]{Pelletier94theprinciple}
Francis~Jeffry Pelletier. 1994.
\newblock The principle of semantic compositionality.
\newblock \emph{Topoi}, 13:11--24.

\bibitem[{Pham et~al.(2018)Pham, Niehues, and Waibel}]{W18-2712}
Ngoc-Quan Pham, Jan Niehues, and Alexander Waibel. 2018.
\newblock \href {http://aclweb.org/anthology/W18-2712} {Towards one-shot
  learning for rare-word translation with external experts}.
\newblock In \emph{Proceedings of the 2nd Workshop on Neural Machine
  Translation and Generation}, pages 100--109. Association for Computational
  Linguistics.

\bibitem[{Sennrich et~al.(2016)Sennrich, Haddow, and
  Birch}]{sennrich-haddow-birch:2016:P16-12}
Rico Sennrich, Barry Haddow, and Alexandra Birch. 2016.
\newblock \href {http://www.aclweb.org/anthology/P16-1162} {Neural machine
  translation of rare words with subword units}.
\newblock In \emph{Proceedings of the 54th Annual Meeting of the Association
  for Computational Linguistics (Volume 1: Long Papers)}, pages 1715--1725,
  Berlin, Germany. Association for Computational Linguistics.

\bibitem[{Snover et~al.(2006)Snover, Dorr, Schwartz, Micciulla, and
  Makhoul}]{Snover06astudy}
Matthew Snover, Bonnie Dorr, Richard Schwartz, Linnea Micciulla, and John
  Makhoul. 2006.
\newblock A study of translation edit rate with targeted human annotation.
\newblock In \emph{In Proceedings of Association for Machine Translation in the
  Americas}, pages 223--231.

\bibitem[{Stanovsky et~al.(2019)Stanovsky, Smith, and
  Zettlemoyer}]{stanovsky-etal-2019-evaluating}
Gabriel Stanovsky, Noah~A. Smith, and Luke Zettlemoyer. 2019.
\newblock \href {https://doi.org/10.18653/v1/P19-1164} {Evaluating gender bias
  in machine translation}.
\newblock In \emph{Proceedings of the 57th Annual Meeting of the Association
  for Computational Linguistics}, pages 1679--1684, Florence, Italy.
  Association for Computational Linguistics.

\bibitem[{Vaswani et~al.(2017)Vaswani, Shazeer, Parmar, Uszkoreit, Jones,
  Gomez, Kaiser, and Polosukhin}]{vaswani2017attention}
Ashish Vaswani, Noam Shazeer, Niki Parmar, Jakob Uszkoreit, Llion Jones,
  Aidan~N Gomez, \L~ukasz Kaiser, and Illia Polosukhin. 2017.
\newblock \href
  {http://papers.nips.cc/paper/7181-attention-is-all-you-need.pdf} {Attention
  is all you need}.
\newblock In I.~Guyon, U.~V. Luxburg, S.~Bengio, H.~Wallach, R.~Fergus,
  S.~Vishwanathan, and R.~Garnett, editors, \emph{Advances in Neural
  Information Processing Systems 30}, pages 5998--6008. Curran Associates, Inc.

\end{thebibliography}
\bibliographystyle{acl_natbib}

\end{document}